\documentclass[11pt]{article}
\usepackage[hidelinks]{hyperref}
\usepackage[authoryear]{natbib}

\usepackage{times}
\usepackage{url}
\usepackage{graphicx}
\usepackage{danudefs}
\usepackage{color}
\usepackage{enumerate}
\usepackage{subcaption}
\usepackage{multirow}
\usepackage{booktabs}
\usepackage{todonotes}

\newcommand{\newcite}{\citet}

\usepackage{amsmath,amsthm,amssymb,fullpage,color,bm}

\theoremstyle{plain}
\newtheorem{lemma}[]{Lemma}

\newcommand{\bmone}{{\bm 1}}
\newcommand{\caC}{\mathcal{C}}
\newcommand{\caD}{\mathcal{D}}
\newcommand{\caF}{\mathcal{F}}
\newcommand{\caN}{\mathcal{N}}

\newcommand{\E}{\mathop{\mathbf{E}}}
\newcommand{\set}[1]{\{#1\}}

\newcommand{\Langle}{\left\langle}
\newcommand{\Rangle}{\right\rangle}

\usepackage{numprint}
\npdecimalsign{.}
\nprounddigits{3}

\begin{document}

\title{Using $k$-way Co-occurrences for Learning Word Embeddings}

\author{Danushka Bollegala\\ University of Liverpool , UK. \\ \texttt{danushka.bollegala@liverpool.ac.uk}\and 
Yuichi Yoshida \\ National Institute of Informatics, Japan. \\ \texttt{yyoshida@nii.ac.jp} \and  Ken-ichi Kawarabayashi \\ National Institute of Informatics, Japan. \\ \texttt{k\_keniti@nii.ac.jp}}
\date{}

\maketitle
\begin{abstract}
Co-occurrences between two words provide useful insights into the semantics of those words.
Consequently, numerous prior work on word embedding learning have used co-occurrences between two words
as the training signal for learning word embeddings.
However, in natural language texts it is common for multiple words to be related and co-occurring in the same context.
We extend the notion of co-occurrences to cover $k(\geq\!\!2)$-way co-occurrences among a set of $k$-words.
Specifically, we prove a theoretical relationship between the joint probability of $k(\geq\!\!2)$ words, and the sum of $\ell_2$
norms of their embeddings. Next, we propose a learning objective motivated by our theoretical result
that utilises $k$-way co-occurrences for learning word embeddings.
Our experimental results show that the derived theoretical relationship does indeed hold empirically, and
despite data sparsity, for some smaller $k$ values, 
$k$-way embeddings perform comparably or better than $2$-way embeddings in a range of tasks.
\end{abstract}

\section{Introduction}

Word co-occurrence statistics are used extensively in a wide-range of NLP tasks for semantic modelling~\cite{Turney:JAIR:2010,Church:ACL:1990}.
As the popular quote from Firth---\emph{you shall know a word by the company it keeps}~\cite{Firth:1957}, the words that co-occur with
a particular word provide useful clues about the semantics of the latter word.
Co-occurrences of a target word with other (context) words in some context such as a fixed-sized window, phrase,
or a sentence have been used for creating word representations~\cite{Mikolov:NAACL:2013,Milkov:2013,Pennington:EMNLP:2014}.
For example, skip-gram with negative sampling (SGNS)~\cite{Milkov:2013} considers the co-occurrences of two words within some local context,
whereas global vector prediction (GloVe)~\cite{Pennington:EMNLP:2014}
 learns word embeddings that can predict the total number of co-occurrences in a corpus.

Unfortunately, much prior work in NLP are limited to the consideration of co-occurrences between two words due to the ease of modelling and data sparseness.
Pairwise co-occurrences can be easily represented using a co-occurrence matrix, whereas co-occurrences involving more than two words would require a higher-order tensor~\cite{Socher:NIPS:2013b}.
Moreover, co-occurrences involving more than three words tend to be sparse even in large corpora~\cite{Turney:JAIR:2012},
requiring compositional approaches for representing their semantics~\cite{zhang-EtAl:2014:EMNLP20145,vandecruys-poibeau-korhonen:2013:NAACL-HLT}.
It remains unknown --  \emph{what statistical properties about words we can learn from $k$-way co-occurrences among words}.
Here, we use the term $k$-way co-occurrence to denote the co-occurrence between $k$-words in some context.

Words do not necessarily appear as pairs in sentences.
By splitting the contexts into pairs of words, we loose the rich contextual information about the nature of the co-occurrences.
For example, consider the following sentences.
\begin{enumerate}[(a)]
\item \emph{John and Anne are friends.}
\item \emph{John and David are friends.}
\item  \emph{Anne and Mary are friends.}
\end{enumerate}
Sentence (a) describes a three-way co-occurrence among (\emph{John}, \emph{Anne}, \emph{friend}), which if
split would result in three two-way co-occurrences: (\emph{John}, \emph{Anne}), (\emph{John}, \emph{friends}), and (\emph{Anne}, \emph{friends}).
On the other hand, Sentences (b) and (c) would collectively produce
the same two two-way co-occurrences (\emph{John}, \emph{friend}) and (\emph{Anne}, \emph{friend}), despite not mentioning
any friendship between \emph{John} and \emph{Anne}.
Therefore, by looking at the three two-way co-occurrences produced by Sentence (a) we cannot unambiguously determine whether \emph{John} and \emph{Anne} are friends.
Therefore, we must retain the three-way co-occurrence (\emph{John}, \emph{Anne}, \emph{friend}) to preserve this information.

Although considering $k$-way co-occurrences is useful for retaining the contextual information, there are several challenges one must
overcome. First, the number of $k$-way co-occurrences tend to be sparse for larger $k$ values.
Such sparse co-occurrence counts might be inadequate for learning reliable and accurate semantic representations.
Second, the unique number of $k$-way co-occurrences grows exponentially with $k$.
This becomes problematic in terms of memory requirements when storing all $k$-way co-occurrences.
A word embedding learning method that considers $k$-way co-occurrences must overcome those two challenges.

In this paper, we make several contributions towards the understanding of $k$-way co-occurrences.
\begin{itemize}
\item We prove a theoretical relation between the joint probability of $k$ words, and the squared sum of $\ell_2$ norms of their embeddings (\S \ref{sec:main-th}).
For this purpose, we extend the work by \newcite{Arora:TACL:2016} for two-way co-occurrences to $k(>2)$-way co-occurrences.

\item Motivated by our theoretical analysis, we propose an objective function
that considers $k$-way co-occurrences for learning word embeddings (\S \ref{sec:train}).
We note that our goal in this paper is \emph{not} to propose novel word embedding learning methods, nor we claim that $k$-way embeddings
produce state-of-the-art results for word embedding learning. 
Nevertheless, we can use word embeddings learnt from $k$-way co-occurrences to empirically evaluate what type of information is captured
by $k$-way co-occurrences.

\item We evaluate the word embeddings created from $k$-way co-occurrences  on multiple benchmark datasets for semantic similarity measurement, analogy detection,
relation classification, and short-text classification (\S \ref{sec:embeds}).
Our experimental results show that, despite data sparsity, for smaller $k$-values such as 3 or 5, $k$-way embeddings outperform $2$-way embeddings.
\end{itemize}

\section{Related Work}

\nocite{Duc:AAAI:2011,Hugo:EMNLP:2010,Noman:2011,Bollegala:JSAI:2007,Hernault:CICLING:2010,Bollegala:IS:2012,Bollegala:IJCNLP:2005,Bollegala_ECAI_2008}

The use of word co-occurrences to learn lexical semantics has a long history in NLP~\cite{Turney:JAIR:2010}.
Counting-based distributional models of semantics, for example, represent a target word by a high dimensional sparse vector
in which the elements correspond to words that co-occur with the target word in some contextual window.
Numerous word association measures such as pointwise mutual information (PMI)~\cite{Church:ACL:1990}, log-likelihood ratio (LLR)~\cite{LLR}, $\chi^{2}$ measure~\cite{chisquare}, etc.
have been proposed to evaluate the strength of the co-occurrences between two words.

On the other hand, prediction-based approaches~\cite{Milkov:2013,Pennington:EMNLP:2014,Collobert:ICML:2008,Mnih:HLBL:NIPS:2008,Huang:ACL:2012} learn low-dimensional
 dense embedding vectors that can be used to accurately predict the co-occurrences between words in some context.
However, most prior work on co-occurrences have been limited to the consideration of two words, whereas
continuous bag-of-words (CBOW)~\cite{Milkov:2013} model is a notable exception because it uses \emph{all} the words in the context of a target word
to predict the occurrence of the target word.
The context can be modelled either as the concatenation or average of the context vectors.
Models that preserve positional information in local contexts have also been proposed~\cite{ling-EtAl:2015:NAACL-HLT}.

Co-occurrences of multiple consecutive words in the form of lexico-syntactic patterns have been successfully applied in tasks that require
modelling of semantic relations between two words.
For example, Latent Relational Analysis (LRA)~\cite{Turney_CL} represents the relations between word-pairs by a co-occurrence matrix where
rows correspond to word-pairs and columns correspond to various lexical patterns that co-occur in some context with the word-pairs.
The elements of this matrix are the co-occurrence counts between the word-pairs and lexical patterns.
However, exact occurrences of $n$-grams tend to be sparse for large $n$ values, resulting in a sparse co-occurrence matrix~\cite{Turney:JAIR:2012}.
LRA uses singular value decomposition (SVD) to reduce the dimensionality, thereby reducing sparseness.

Despite the extensive applications of word co-occurrences in NLP, theoretical relationships between co-occurrence statistics
and semantic representations have been less understood.
\newcite{Hashimoto:TACL:2016} show that word embedding learning can be seen as a problem of metric recovery from log co-occurrences between words in a large corpus.
\newcite{Arora:TACL:2016} show that log joint probability between two words is proportional to the squared sum of the $\ell_{2}$ norms of
their embeddings.
However, both those work are limited to two-way co-occurrences (i.e. $k=2$ case).
In contrast, our work can be seen as extending this analysis to $k > 2$ case.
In particular, we show that under the same assumptions made by \newcite{Arora:TACL:2016}, the log joint probability of a set of $k$ co-occurring words is proportional to
the squared sum of $\ell_{2}$ norms of their embeddings.

Averaging word embeddings to represent sentences or phrases has found to be a simple yet an accurate method~\cite{Arora:ICLR:2017,Kenter:ACL:2016} that has reported comparable performances to more complex models that consider the ordering of words~\cite{Kiros:2015}.
For example, \newcite{Arora:ICLR:2017} compute sentence embeddings as the linearly weighted sum of the constituent word embeddings, where the weights
are computed using unigram probabilities, whereas \newcite{Rei:2016} propose a task-specific supervised approach for learning the combination weights.
\newcite{Kenter:ACL:2016} learn word embeddings such that when averaged produce accurate sentence embeddings.
Such prior work hint at the existence of a relationship between the summation of the word embeddings, and the semantics of the sentence that
contains those words.
However, to the best of our knowledge, a theoretical connection between $k$-way co-occurrences and word embeddings has not been established before.

\section{$k$-way word co-occurrences}
\label{sec:main-th}

Our analysis is based on the \emph{random walk model} of text generation proposed by \newcite{Arora:TACL:2016}
Let $\cV$ be the vocabulary of words.
Then, the $t$-th word $w_{t} \in \cV$ is produced at step $t$ by a random walk driven by a discourse vector $\vec{c}_{t} \in \R^{d}$.
Here, $d$ is the dimensionality of the embedding space and coordinates of $\vec{c}_{t}$ represent what is being talked about.
Moreover, each word $w \in \cV$ is represented by a vector (embedding) $\vec{w} \in \R^{d}$.
Under this model, the probability of emitting $w \in \cV$ at time $t$, given $\vec{c}_{t}$ given by~\eqref{eq:model}.
\begin{equation}
 \label{eq:model}
 \textrm{Pr}[\text{emitting $w$ at time $t$} \mid \vec{c}_{t}] \propto \exp \left( \vec{c}_{t}\T\vec{w} \right)
\end{equation}
Here, a \emph{slow} random work is assumed where $\vec{c}_{t+1}$ can be obtained from $\vec{c}_{t}$ by adding a small
random displacement vector such that nearby words are generated under similar discourses.
More specificaly, we assume that $\|\vec{c}_{t+1}-\vec{c}_t\|_2 \leq \epsilon_2/\sqrt{d}$ for some small $\epsilon_2>0$.
The stationary distribution $\cC$ of the random walk is assumed to be uniform over the unit sphere.
For such a random walk, \newcite{Arora:TACL:2016} prove the following Lemma.
\begin{lemma}[\textbf{Concentration of Partition functions} Lemma 2.1 of~\cite{Arora:TACL:2016}]\label{lem:2.1-of-Arora}
  If the word embedding vectors satisfy the Bayesian prior $\vec{v} = s \hat{\vec{v}}$, where $\vec{\hat{v}}$ is from the spherical Gaussian distribution, and $s$ is a scalar random variable, which is always bounded by a constant, then the entire ensemble of word vectors satisfies that
  \begin{align}
    \Pr_{c \sim \caC}[(1-\epsilon_z) Z \leq Z_c \leq (1+\epsilon_z)Z]\geq 1-\delta, \label{eq:2.1-of-Arora}
  \end{align}
  for $\epsilon_z = O (1/\sqrt{n})$, and $\delta = \exp(-\Omega(\log^2 n))$,
  where $n \geq d$ is the number of words and $Z_{c}$ is the partition function for $c$ given by $\sum_{w \in \cV}  \exp \left( \vec{w}\T\vec{c} \right)$.

\end{lemma}
 Lemma 1 states that the partition function concentrates around a constant value $Z$ for all $c$ with high probability.

For $d$ dimensional word embeddings, the relationship between the $\ell_{2}$ norm of word embeddings $\vec{w}_{i}$, $\norm{\vec{w}_{i}}_{2}$, and the joint probability of the words, $p(w_{1}, \ldots, w_{k})$ is given by the following theorem: 
\begin{theorem}
\label{the:joint-probability}
  Suppose the word vectors satisfy~\eqref{eq:2.1-of-Arora}.
  Then, we have
  \begin{align}
  \label{eq:joint-prob}
    \log p(w_1,\ldots,w_k) = \frac{\norm{\sum_{i=1}^k\vec{w}_i}_{2}^2}{2d} - k \log Z \pm \epsilon.
  \end{align}
  for $\epsilon = O(k \epsilon_z) + \widetilde{O}(1/d) + O(k^2 \epsilon_2)$, where
{\small
\begin{align}
\small
 \label{eq:Z}
 Z = \sum_{(w_{1}, \ldots, w_{k}) \in \cV^k} \sum_{c \in \cC} \exp\left( \sum_{i=1}^k \vec{w}_{i}^\top\vec{c} \right).
\end{align}}
\end{theorem}
Note that the normalising constant (partitioning function) $Z$ given by~\eqref{eq:Z} is independent of the co-occurrences.

Proof of \autoref{the:joint-probability} is given in the Appendix.
In particular, for $k = 1$ and $2$, \autoref{the:joint-probability} reduces to the relationships proved by \newcite{Arora:TACL:2016}.
Typically the $\ell_{2}$ norm of $d$ dimensional word vectors is in the order of $\sqrt{d}$, implying that
 the order of the squared $\ell_2$ norm of $\sum_{i=1}^k \vec{w}_{i}$ is $\O(d)$.
 Consequently, the noise level $\O(\epsilon)$ is significantly smaller compared to the first term in the left hand side.
 Later in \S~\ref{sec:empirical}, we empirically verify the relationship stated in \autoref{the:joint-probability}
 and the concentration properties of the partitioning function for $k$-way co-occurrences.

\section{Learning $k$-way Word Embeddings}
\label{sec:train}

In this Section, we propose a training objective that considers $k$-way co-occurrences using the relationship given by \autoref{the:joint-probability}.
By minimising the proposed objective we can obtain word embeddings that consider $k$-way co-occurrences among words.
The word embeddings derived in this manner serve as a litmus test for empirically evaluating the validity of \autoref{the:joint-probability}.

Let us denote the $k$-way co-occurrence $(w_{1}, \ldots, w_{k}) = w_{1}^{k}$, and its frequency in a corpus by $h(w_{1}^{k})$.
The joint probability $p(w_{1}^{k})$ of such a $k$-way co-occurrence is given by~\eqref{eq:joint-prob}.
Although successive samples from a random walk are not independent, if we assume the random walk to mix fairly quickly
(i.e.~mixing time related to the logarithm of the vocabulary size), then the distribution of $h(w_{1}^{k})$ can be approximated by
a multinomial distribution $\textsf{Mul}\left(\tilde{L}_{k}, \{p(w_{1}^{k})\}\right)$, where $\tilde{L}_k = \sum_{w_{1}^{k} \in \cG_{k}} h(w_{1}^{k})$
and $\cG_{k}$ is the set of all $k$-way co-occurrences.
Under this approximation, Theorem~\ref{the:embed} provides an objective for learning word embeddings from $k$-way co-occurrences.

\begin{theorem}
\label{the:embed}

The set of word embeddings $\{\vec{w}_{i}\}$ that minimise the objective given by~\eqref{eq:obj} maximises the log-likelihood of
$k$-way co-occurrences given by~\eqref{eq:likelihood}. Here, $C$ is a constant independent of the word embeddings.
{\small
\begin{align}
\label{eq:obj}
\sum_{w_{1}^{k} \in \cG_{k}} h(w_{1}^{k}) \left( \log(h(w_{1}^{k})) - \norm{\sum_{i=1}^{k} \vec{w}_{i}}_{2}^{2} + C \right)^{2}
\end{align}}
{\small
\begin{align}
\label{eq:likelihood}
l = \log \left( \prod_{w_{1}^{k} \in \cG_{k}}{p(w_{1}^{k})}^{h(w_{1}^{k})} \right)
\end{align}}
\end{theorem}

\begin{proof}
The log-likelihood term can we written as
{\small
\begin{align}
\label{eq:LL}
 l = \sum_{w_{1}^{k} \in \cG_{k}} h(w_{1}^{k}) \log p(w_{1}^{k}) .
\end{align}}
The expected count of a $k$-way co-occurrence $w_{1}^{k}$ can be estimated as $\tilde{L}_k p(w_{1}^{k})$.
We then define the log-ratio between the expected and actual $k$-way co-occurrence counts $\Delta_{w_{1}^{k}}$ as
{\small
\begin{align}
 \label{eq:LLR}
 \Delta_{w_{1}^{k}} = \log \left( \frac{\tilde{L}_k p(w_{1}^{k})}{h(w_{1}^{k})} \right) .
\end{align}}
Substituting for $p(w_{1}^{k})$ from~\eqref{eq:LLR} in~\eqref{eq:LL} we obtain
{\small
\begin{align}
\label{eq:LL2}
l = \sum_{w_{1}^{k} \in \cG_{k}} h(w_{1}^{k}) \left(  \Delta_{w_{1}^{k}} - \log \tilde{L}_k + \log h(w_{1}^{k}) \right) .
\end{align}}
Representing the terms independent from the embeddings $\vec{w}_{k}$ by $C$ we  can re-write~\eqref{eq:LL2} as
{\small
\begin{align}
 \label{eq:LL3}
 l = C +  \sum_{w_{1}^{k} \in \cG_{k}} h(w_{1}^{k}) \Delta_{w_{1}^{k}} .
\end{align}}
Because $p(w_{1}^{k})$ represents a joint probability distribution over $k$-way co-occurrences $w_{1}^{k}$ we have
{\small
\begin{align}
\label{eq:L1}
 \tilde{L}_k = \sum_{w_{1}^{k} \in \cG_{k}} \tilde{L}_k p(w_{1}^{k})  .
\end{align}}
Substituting~\eqref{eq:LLR} in~\eqref{eq:L1} we obtain
{\small
\begin{align}
 \label{eq:L2}
  \tilde{L}_k = \sum_{w_{1}^{k} \in \cG_{k}} h(w_{1}^{k}) \exp \left( \Delta_{w_{1}^{k}} \right)  .
\end{align}}
When $x$ is small, from Taylor expansion $\exp(x) \approx 1 + x + x^{2}/2$ we have
{\small
\begin{align}
 \label{eq:L3}
 \tilde{L}_k \approx \sum_{w_{1}^{k} \in \cG_{k}} h(w_{1}^{k}) \left( 1 + \Delta_{w_{1}^{k}} + \frac{\Delta^{2}_{w_{1}^{k}}}{2} \right)
\end{align}}
Although this Taylor expansion has an approximation error of $\mathcal{O}(x^{3})$, for large $h(w_{1}^{k})$ values,
expected counts approach actual counts resulting in $\Delta_{k}$ values closer to 0 according to~\eqref{eq:LLR}.
On the other hand, word co-occurrence counts approximately follow a power-law distribution~\cite{Pennington:EMNLP:2014}.
Therefore, contributions to the objective function by $\Delta_{w_{1}^{k}}$ terms corresponding to smaller $h(w_{1}^{k})$ can be ignored in practice.
Then, by definition we have
{\small
\begin{align}
 \label{eq:defL}
 \tilde{L}_k =  \sum_{w_{1}^{k} \in \cG_{k}} h(w_{1}^{k}) .
\end{align}}
By substituting~\eqref{eq:defL} in~\eqref{eq:L3} we obtain
{\small
\begin{align}
\label{eq:L4}
  \sum_{w_{1}^{k} \in \cG_{k}} h(w_{1}^{k}) \Delta_{w_{1}^{k}} \approx -\frac{1}{2}  \sum_{w_{1}^{k} \in \cG_{k}} h(w_{1}^{k}) \Delta^{2}_{w_{1}^{k}} .
\end{align}}
From~\eqref{eq:LL3} and~\eqref{eq:L4} we obtain
{\small
\begin{align}
 \label{eq:L5}
 l = C - \frac{1}{2}  \sum_{w_{1}^{k} \in \cG_{k}} h(w_{1}^{k}) \Delta^{2}_{w_{1}^{k}} .
\end{align}}
Therefore, minimisation of  $\sum_{w_{1}^{k} \in \cG_{k}} h(w_{1}^{k}) \Delta^{2}_{w_{1}^{k}}$ corresponds to the maximisation of the log-likelihood.
\end{proof}

Minimising the objective~\eqref{eq:obj} with respect to $\vec{w}_{i}$ and $C$ produces word embeddings that capture the relationships
in $k$-way co-occurrences of words in a corpus. Down-weighting very frequent co-occurrences of words has shown to be effective
in prior work. This can be easily incorporated into the objective function~\eqref{eq:obj} by replacing $h(w_{1}^{k})$ by a truncated version such as
$\min(h(w_{1}^{k}), \theta_{k})$, where $\theta$ is a cut-off threshold, where we set $\theta = 100$ following prior work.
We find the word embeddings $\vec{w}_{i}$ for a set of $k$-way co-occurrences $\cG_{k}$ and the parameter $C_{k}$, by
computing the partial derivative of the objective given by \autoref{eq:obj} w.r.t.~those parameters, and applying Stochastic Gradient Descent (SGD)
with learning rate updated using AdaGrad~\cite{Duchi:JMLR:2011}. The initial learning rate is set to $0.01$ in all experiments.
We refer to the word embeddings learnt by optimising \eqref{eq:obj} as \textbf{$k$-way embeddings}.

\section{Experiments}
\label{sec:exp}

We pre-processed a January 2017 dump of English Wikipedia using a Perl script\footnote{\url{http://mattmahoney.net/dc/textdata.html}}
 and used as our corpus (contains ca. 4.6B tokens).
 We select unigrams occurring at least $1000$ times in this corpus amounting to a vocabulary of size $73,954$.
 Although it is possible to apply the concept of $k$-way co-occurrences to $n$-grams of any length $n$,
 for the simplicity we limit the analysis to co-occurrences among unigrams.
 Extracting $k$-way co-occurrences from a large corpus is challenging because of the large number of unique and sparse $k$-way co-occurrences.
Note that $k$-way co-occurrences are however less sparse and less diverse compared to $k$-grams because the ordering of words is ignored in a $k$-way co-occurrence. Following the Apriori algorithm~\cite{Apriori} for extracting frequent itemsets of a particular length with a pre-defined support,
we extract $k$-way co-occurrences that occur at least $1000$ times in the corpus within a $10$ word window.

Specifically, we select all $(k-1)$-way co-occurrences that occur at least $1000$ times and grow them by appending the selected unigrams
(also occurring at least $1000$ times in the corpus). We then check whether all subsets of length $(k-1)$ of a candidate $k$-way co-occurrence
appear in the set of frequent $(k-1)$-way co-occurrences.
If this requirement is satisfied, then it follows from the apriori property that the generated $k$-way co-occurrence must have a minimum support of $1000$.
Following this procedure we extract $k$-way co-occurrences for $k = 2, 3, 4$, and $5$ as shown in \autoref{tbl:counts}.

\begin{table}[t]
\small
\centering
\begin{tabular}{ll}\toprule
$k$ & no.~of $k$-way co-occurrences \\ \midrule
2 & 257,508,996 \\
3 & 394,670,208 \\
4 & 111,119,411 \\
5 & 14,495,659 \\ \bottomrule
\end{tabular}
\caption{The number of unique $k$-way co-occurrence with support $1000$.}
\label{tbl:counts}
\end{table}

\subsection{Empirical Verification of the Model}
\label{sec:empirical}

\begin{figure*}[t]
    \begin{subfigure}[t]{0.25\textwidth}
        \centering
        \includegraphics[width=48mm]{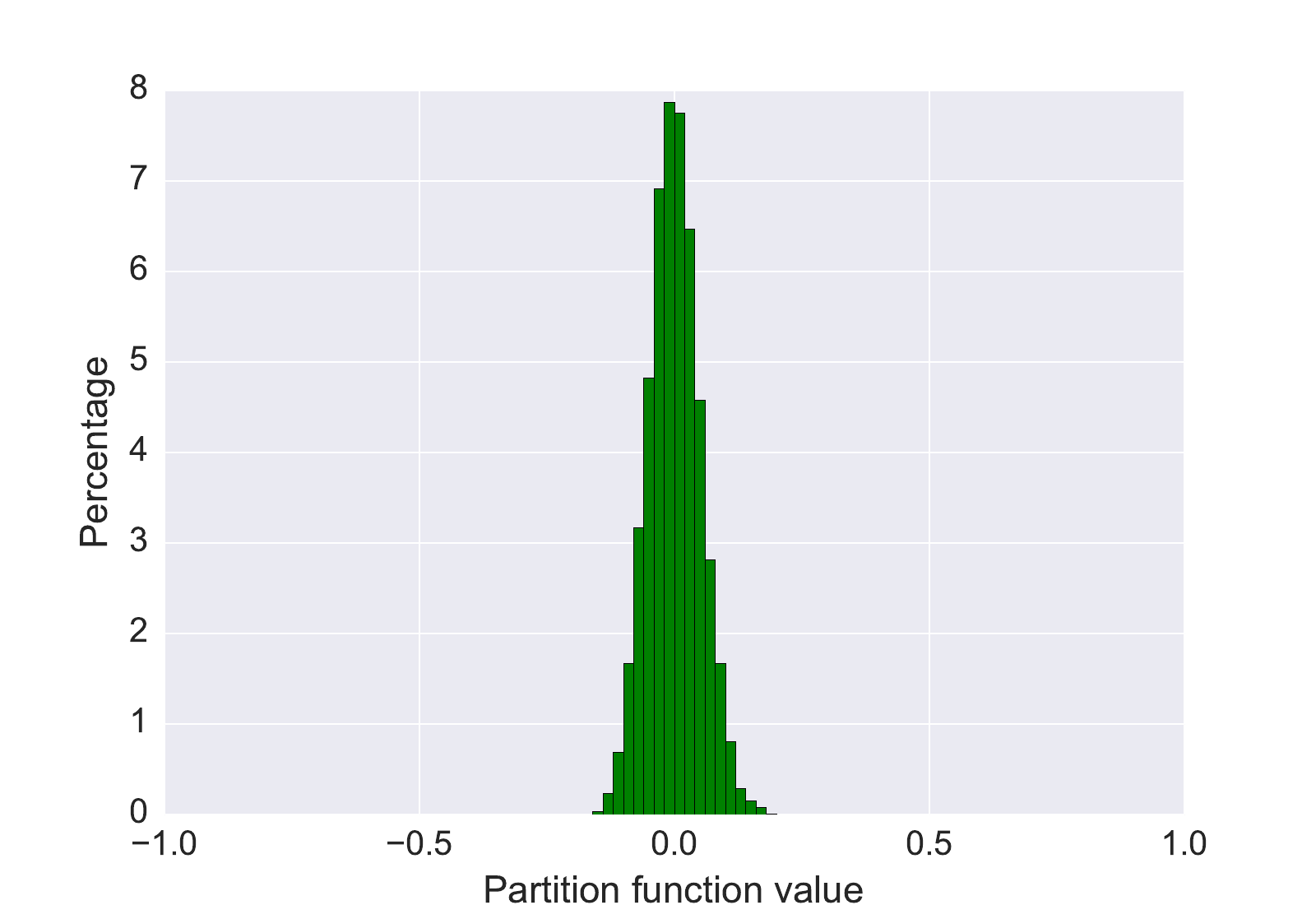}
        \caption{2-way co-occurrences}
    \end{subfigure}%
     \hfill
    \begin{subfigure}[t]{0.25\textwidth}
        \centering
        \includegraphics[width=48mm]{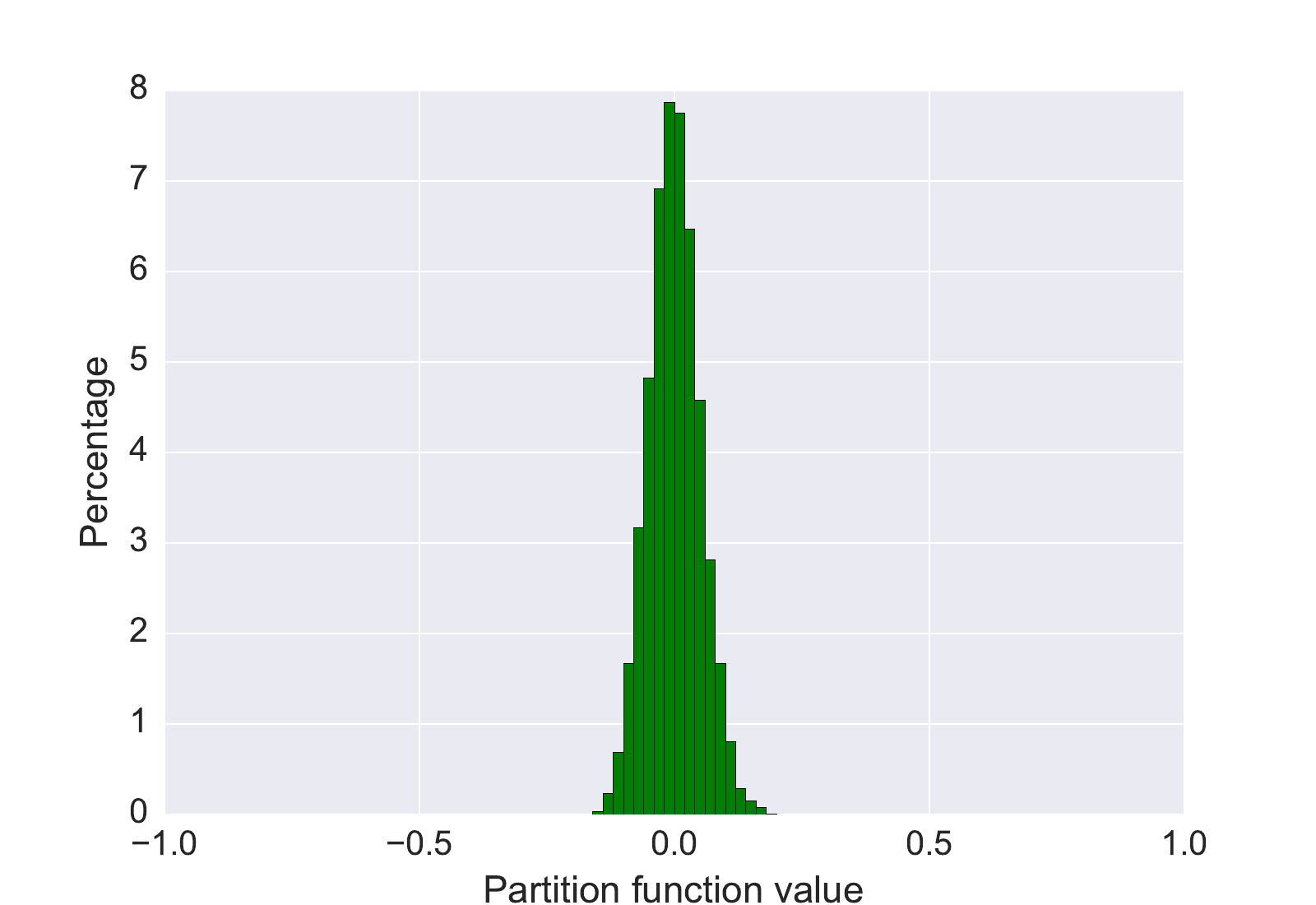}
        \caption{3-way co-occurrences}
    \end{subfigure}
\hfill
     \begin{subfigure}[t]{0.25\textwidth}
        \centering
        \includegraphics[width=48mm]{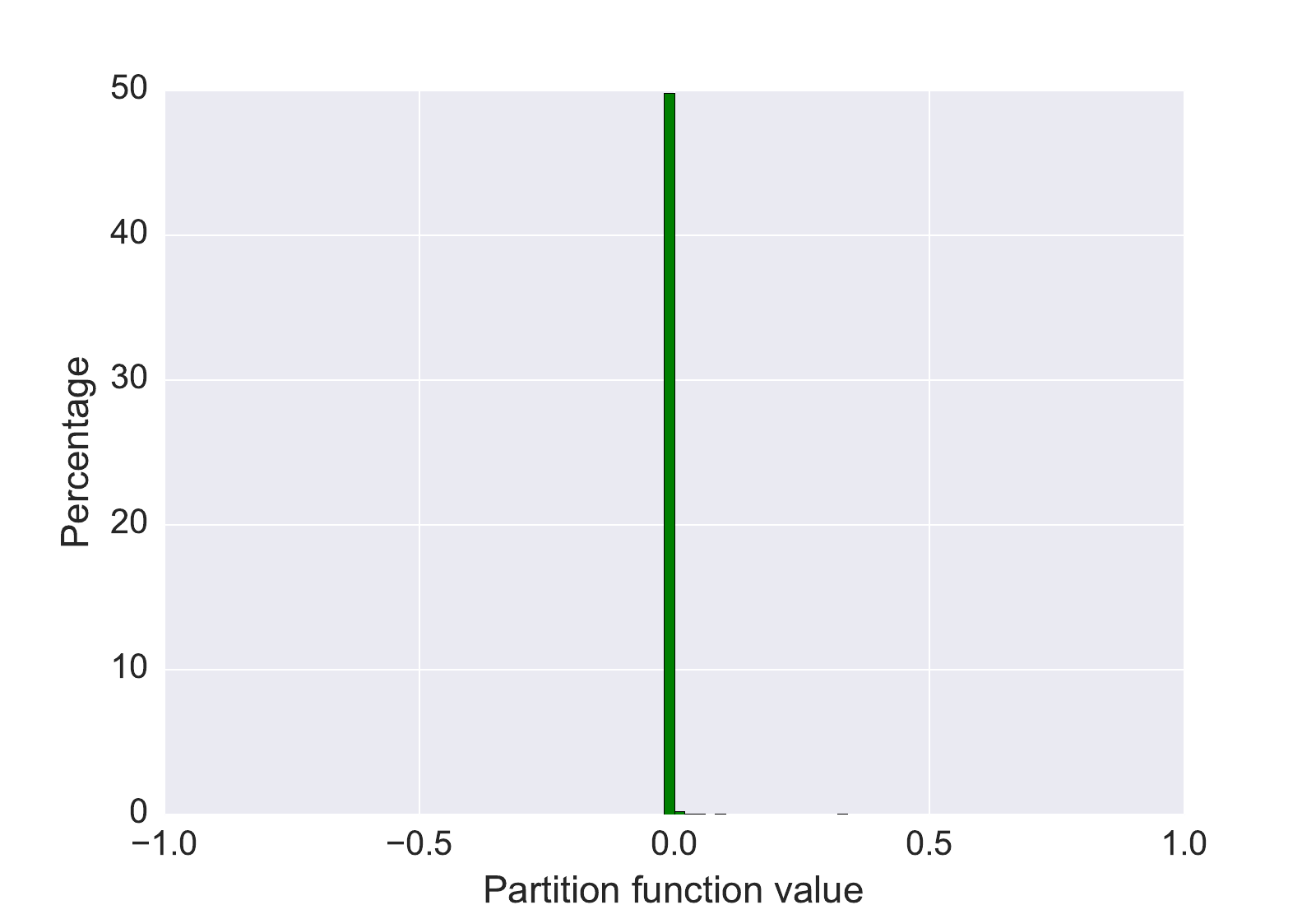}
        \caption{4-way co-occurrences}
    \end{subfigure}%
\hfill
    \begin{subfigure}[t]{0.24\textwidth}
        \centering
        \includegraphics[width=48mm]{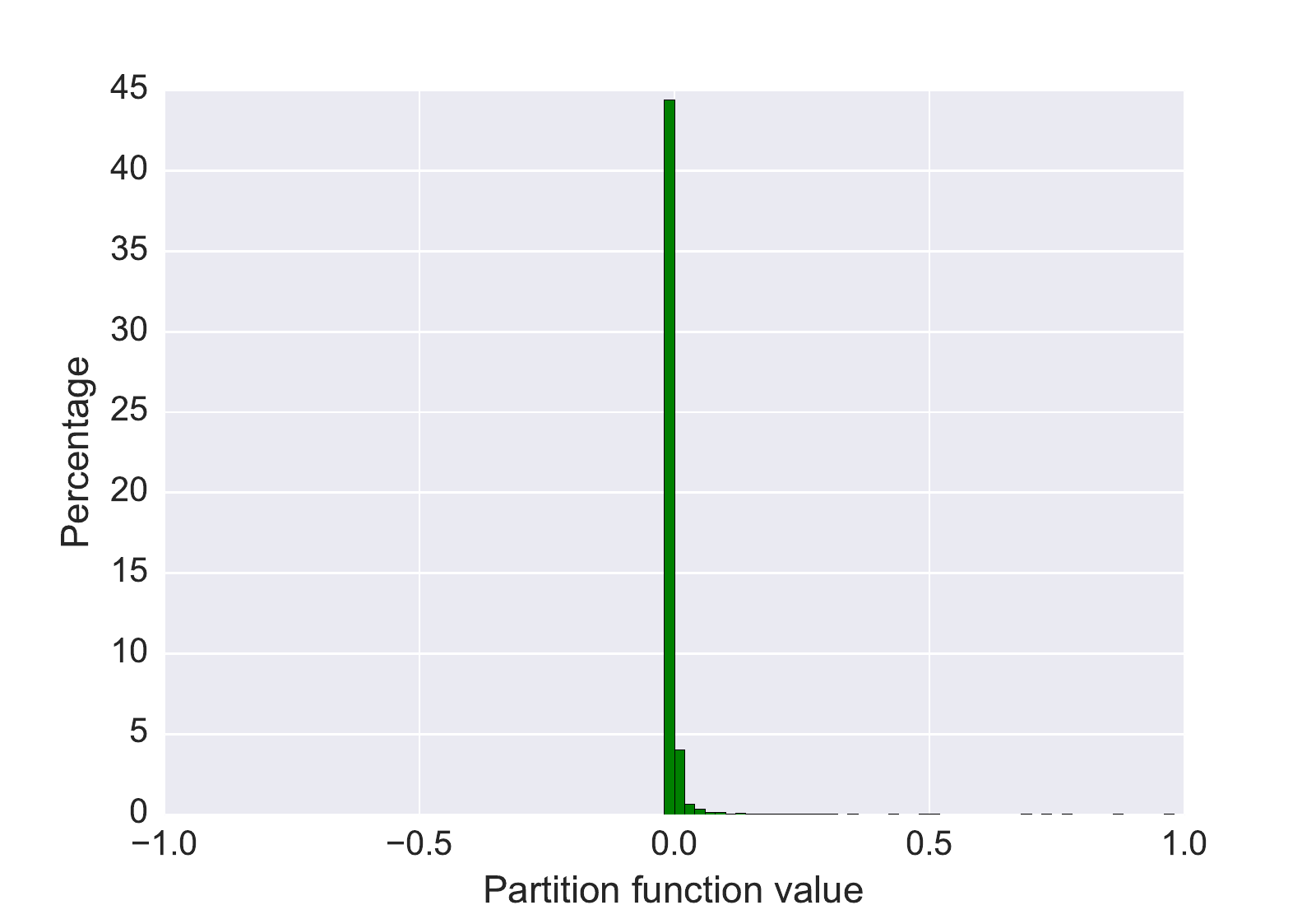}
        \caption{5-way co-occurrences}
    \end{subfigure}
    \caption{Histogram of the partitioning function for randomly chosen $10,000$ context vectors.}
    \label{fig:partition}
\end{figure*}

Our proof of \autoref{the:joint-probability} requires the condition used in Lemma~\autoref{lem:2.1-of-Arora}, which states that the partition function
given by \eqref{eq:Z} must concentrate within a small range for any $k$.
Although such concentration properties for $2$-way co-occurrences have been reported before,
it remains unknown whether this property holds for $k(>\!\!2)$-way co-occurrences.
To test this property empirically, we uniformly randomly generate $10^{5}$ vectors $\vec{c}$ ($\ell_{2}$ normalised to unit length) and compute the histogram of the partition function values as shown in \autoref{fig:partition} for $d = 300$ dimensional embeddings.
We standardise the histogram to zero mean and unit variance for the ease of comparisons.
From \autoref{fig:partition}, we see that the partition function concentrates around the mean for all $k$-values.
Interestingly, the concentration is stronger for higher $k(>\!\!3$) values.
Because we compute the sum of the embeddings of individual words in \eqref{eq:Z}, from the law of large numbers
it follows that the summation converges towards the mean when we have more terms in the $k$-way co-occurrence.
This result shows that the assumption on which \autoref{the:joint-probability} is based (i.e.~concentration of the partition function
for arbitrary $k$-way co-occurrences), is empirically justified.

Next, to empirically verify the correctness of \autoref{the:joint-probability},
we learn $d=300$ dimensional $k$-way embeddings for each $k$ value in range $[2,5]$ separately , and measure the Spearman correlation between $\log p(w_{1}, \ldots, w_{k})$ and $\norm{\sum_{i=1}^{k} \vec{w}_{i}}_{2}^{2}$ for a randomly selected $10^{6}$ $k$-way co-occurrences.
If \eqref{eq:joint-prob} is correct, then we would expect a linear relationship (demonstrated by a high positive correlation) between the two sets of values for a fixed $k$.

\begin{figure*}[t]
    \centering
    \begin{subfigure}[t]{0.5\textwidth}
        \centering
        \includegraphics[height=50mm]{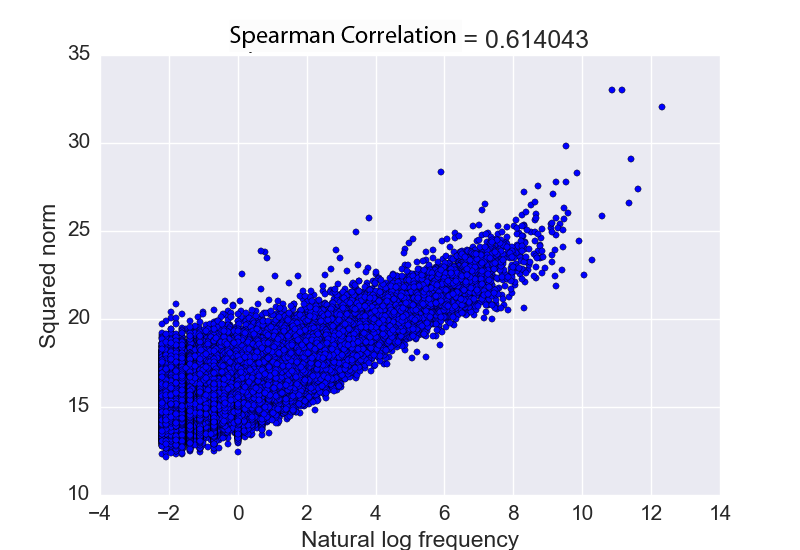}
        \caption{2-way co-occurrences}
    \end{subfigure}%
     \hfill
    \begin{subfigure}[t]{0.5\textwidth}
        \centering
        \includegraphics[height=50mm]{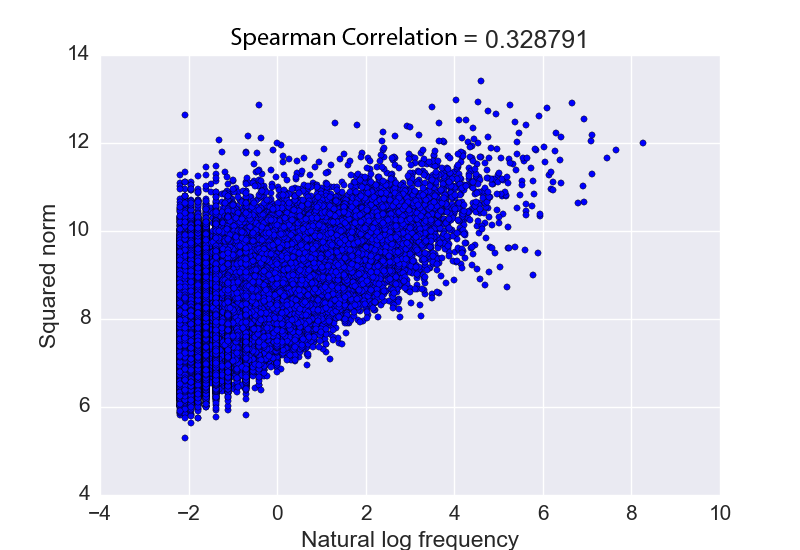}
        \caption{3-way co-occurrences}
    \end{subfigure}
     \vskip\baselineskip
     \begin{subfigure}[t]{0.5\textwidth}
        \centering
        \includegraphics[height=50mm]{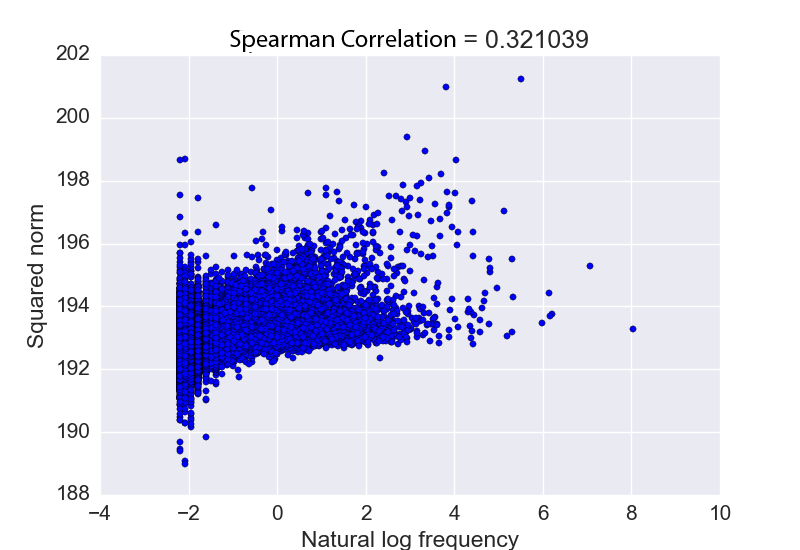}
        \caption{4-way co-occurrences}
    \end{subfigure}%
     \hfill
    \begin{subfigure}[t]{0.5\textwidth}
        \centering
        \includegraphics[height=50mm]{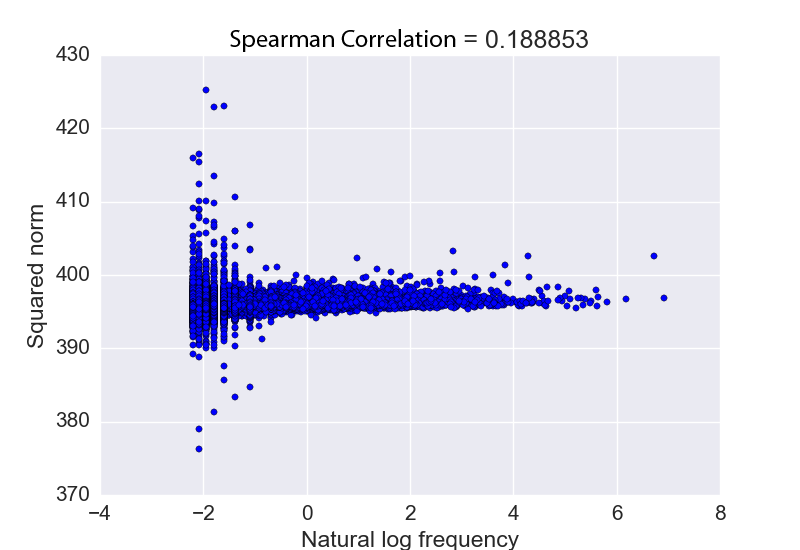}
        \caption{5-way co-occurrences}
    \end{subfigure}
    \caption{Correlation between the squared $\ell_{2}$ norms of the sum of the $k$-way embeddings and the natural log frequency of the corresponding $k$-way co-occurrences are shown for different $k$ values.}
    \label{fig:correlation}
\end{figure*}

\autoref{fig:correlation} shows the correlation plots for $k=2, 3, 4$, and $5$.
From \autoref{fig:correlation} we see that there exist such a positive correlation in all four cases.
However, the value of the correlation drops when we increase $k$ as a result of the sparseness of $k$-way co-occurrences for larger $k$ values.
Although due to the limited availability of space we show results only for $d = 300$ embeddings, the above-mentioned trends could be observed across a wide range of dimensionalities ($d \in [50, 1000]$) in our experiments.

\subsection{Evaluation of Word Embeddings}
\label{sec:embeds}

\begin{table*}
\small
\centering
\begin{tabular}{l l l l l l l l l l l l l l }\toprule
$k$ & RG & MC & WS & RW & SCWS & MEN & SL & SE & DV & TR & MR & CR & SUBJ \\ \midrule
$2$ & \textbf{78.63} & 79.17 & 59.68 & 41.53 & \textbf{57.09} & 70.42 & \textbf{34.76} & 37.21 & 75.34 & 72.43 & 68.38 & 79.19 & 82.20 \\
$\leq3$ & 77.51 & \textbf{79.92} & 59.61 & \textbf{41.58} & 56.69 & $\mathbf{70.92}^{*}$ & 34.65 &  \textbf{37.42} & $\mathbf{75.96}^{*}$ & $\mathbf{72.92}^{*}$ & \textbf{68.71} & $\mathbf{79.52}^{*}$ & 82.35 \\
$\leq4$ & 75.85 & 72.66 & 59.75 & 41.23 & 56.74 & 70.32 & 34.51 & 37.01 & 74.92 & 72.37 & 67.87 & 78.18 & 82.25\\
$\leq5$ & 75.19 & 74.63 & $\mathbf{60.54}^{*}$ & 40.84 & 56.92 & 70.50 & 34.67 & 37.21 & 74.76 & 72.21 & 68.48 & 77.18 & $\mathbf{82.60}^{*}$\\ \bottomrule
\end{tabular}
\caption{The results on word similarity, analogy, relation classification and short-text classification tasks reported by the word embeddings learnt using $k$-way co-occurrences for different $k$ values.}
\label{tbl:res}
\end{table*}

We re-emphasise here that our goal in this paper is \emph{not} to propose novel word embedding learning methods but to extend the notion of $2$-way
co-occurrences to $k$-way co-occurrences.
Unfortunately all existing word embedding learning methods use only $2$-way co-occurrence information for learning.
Moreover, direct comparisons against different word embedding learning methods that use only $2$-way co-occurrences are meaningless here because
the performances of those pre-trained embeddings will depend on numerous factors such as the training corpora, co-occurrence window size, word association measures,
objective function being optimised, and the optimisation methods.
Nevertheless, by evaluating the $k$-way embeddings learnt for different $k$ values using the same resources,
we can empirically evaluate the amount of information captured by $k$-way co-occurrences.

For this purpose, we use four tasks that have been used previously for evaluating word embeddings.
\begin{description}
\item[Semantic similarity measurement:] We measure the similarity between two words as the cosine similarity between the corresponding
embeddings, and measure the Spearman correlation coefficient against the human similarity ratings. We use
\newcite{RG} (\textbf{RG}, 65 word-pairs), \newcite{MC} (\textbf{MC}, 30 word-pairs),
rare words dataset (\textbf{RW}, 2034 word-pairs)~\cite{Luong:CoNLL:2013},
Stanford's contextual word similarities (\textbf{SCWS}, 2023 word-pairs)~\cite{Huang:ACL:2012},
the \textbf{MEN} dataset (3000 word-pairs)~\cite{MEN}, and the SimLex \textbf{SL} dataset~\cite{SimLex} (999 word-pairs).

\item[Word analogy detection:] Using the CosAdd method, we solve word-analogy questions in
 the SemEval (\textbf{SE})  dataset~\cite{SemEavl2012:Task2}.
Specifically, for three given words $a$, $b$ and $c$, we find a fourth word $d$ that correctly answers the question
\emph{$a$ to $b$ is $c$ to what?} such that the cosine similarity between the two vectors $(\vec{b} - \vec{a} + \vec{c})$ and $\vec{d}$
is maximised.

\item[Relation classification:] We use the \textsc{DiffVec} \textbf{DV}~\cite{Vylomova:ACL:2016} dataset containing 12,458 triples
of the form $(\textrm{relation}, \textrm{word}_{1}, \textrm{word}_{2})$ covering 15 relation types. We train a 1-nearest neighbour classifier,
where for each target tuple we measure the cosine similarity between the vector offset for its two word embeddings, and
those of the remaining tuples in the dataset.
If the top ranked tuple has the same relation as the target tuple, then it is considered to be a correct match.
We compute the (micro-averaged) classification accuracy over the entire dataset as the evaluation measure.

\item[Short-text classification:]  We use four binary short-text classification datasets: Stanford sentiment treebank (\textbf{TR})\footnote{\url{http://nlp.stanford.edu/sentiment/treebank.html}} (903 positive test instances and 903 negative test instances),
movie reviews dataset (\textbf{MR})~\cite{Pang:ACL:2005} (5331 positive instances and 5331 negative instances),
customer reviews dataset  (\textbf{CR})~\cite{Hu:KDD:2004} (925 positive instances and 569 negative instances), and 
the subjectivity dataset (\textbf{SUBJ})~\cite{Pang+Lee:04a} (5000 positive instances and 5000 negative instances).
Each review is represented as a bag-of-words and we compute the centroid of the embeddings for each bag to represent the review.
Next, we train a binary logistic regression classifier using the train portion of each dataset, and evaluate the classification accuracy using the
corresponding test portion.
\end{description}
Statistical significance at $p < 0.05$ level is evaluated for correlation coefficients and classification accuracies using respectively Fisher transformation and
Clopper-Pearson confidence intervals.

Learning $k$-way embeddings from $k$-way co-occurrences for a single $k$ value results in poor performance because of data sparseness.
To overcome this issue we use all co-occurrences equal or below a given $k$ value when computing $k$-way embeddings for a given $k$.
Training is done in an iterative manner where we randomly initialise word embeddings when training $k=2$-way embeddings, and
subsequently use $(k-1)$-way embeddings as the initial values for training $k$-way embeddings.
The performances reported by $300$ dimensional embeddings are shown in  \autoref{tbl:res},
where best performance in each task is shown in bold and statistical significance over $2$-way embeddings is indicated by an asterisk.

From \autoref{tbl:res}, we see that for most of the tasks the best performance is reported by $k(\geq2)$-way embeddings and not $k=2$-way embeddings.
In some of the larger datasets, the performances reported by $k\leq3$ (for \textbf{MEN}, \textbf{DV}, and \textbf{CR}) and $k\leq5$ way embeddings (for \textbf{WS} and \textbf{SUBJ}) are significantly better than that by the $2$-way embeddings.
This result supports our claim that $k(>2)$-way co-occurrences should be used in addition to $2$-way co-occurrences when learning word embeddings.

Prior work on relational similarity measurement have shown that the co-occurrence context between two words provide useful clues regarding the semantic relations that exist between those words.
For example, the the phrase \emph{is a large} in the context \emph{\textbf{Ostrich} is a large \textbf{bird}} indicates a hypernymic relation between \emph{ostrich} and \emph{bird}.
The two datasets \textbf{SE} and \textbf{DV} evaluate word embeddings for their ability to represent semantic relations between two words.
Interestingly, we see that $k \leq 3$ embeddings perform best on those two datasets.

Text classification tasks require us to understand not only the meaning of individual words but also the overall topic in the text.
For example, in a product review individual words might have both positive and negative sentiments but for different aspects of the product.
Consequently, we see that $k \leq 3$ embeddings consistently outperform $k = 2$ embeddings on all short-text classification tasks.
By consider all co-occurrences for $k \leq 5$ we see that we obtain the best performance on the \textbf{SUBJ} dataset.

For the word similarity benchmarks, which evaluate the similarity between two words, we see that $2$-way co-occurrences are sufficient to obtain the best results in most cases. A notable exception is \textbf{WS} dataset, which has a high portion of related words than datasets such as \textbf{MEN} or \textbf{SL}.
Because related words can co-occur in broader contextual window and with various words, considering a $k \leq 5$ way co-occurrences seem to be effective.

\subsection{Qualitative Evaluation}
\label{sec:quality}

Our quantitative experiments revealed that 3-way embeddings are particularly better than 2-way embeddings in multiple tasks.
To qualitatively evaluate the difference between 2-way and 3-way embeddings, we conduct the following experiment.

First, we combine all word pairs in semantic similarity benchmarks to create a dataset containing 8483 word pairs with human similarity ratings. We normalise the human similarity ratings in each dataset separately to $[0,1]$ range by subtracting the minimum rating and dividing by the difference between maximum and minimum ratings. The purpose of this normalisation is to make the ratings in different benchmark datasets comparable. Next, we compute the cosine similarity between the two words in each word pair using 2-way and 3-way embeddings separately. We then select word pairs where the difference between the two predicted similarity scores are significantly greater than one standard deviation point. This process yields $911$ word pairs, which we manually inspect and classify into several categories.

 \autoref{tbl:qual} shows some randomly selected word pairs with their predicted similarity scores scaled to 0.5 means and 1.0 variance, and human ratings given in the original benchmark dataset in which the word pair appears. 
 We found that 2-way embeddings assign high similarity scores for many unrelated word pairs, whereas by using 3-way embeddings we are able to reduce the similarity scores assigned to such unrelated word pairs. 
 Words such as \emph{giraffe}, \emph{car} and \emph{happy} are highly frequent and co-occur with many different words.
Under 2-way embeddings, any word that co-occur with a target word will provide a semantic attribute to the target word.
Therefore, unrelated word pairs where at least one word is frequent are likely to obtain relatively higher similarity score under 2-way embeddings.

We see that the similarity between two words in a collocation are overly estimated by 2-way embeddings.
The two words forming a collocation are not necessarily semantically similar. For example, \emph{movie} and \emph{star} do not share
many attributes in common. 3-way embeddings correctly assigns lower similarity scores for such words because many other words
co-occur with a particular collocation in different contexts.

We observed that 2-way embeddings assign high similarity scores for a large number of antonym pairs.
Prior work on distributional methods of word representations have shown that it is difficult to discriminate between antonyms and synonyms using their word distributions~\citep{Mohammad:CL:2013}.
\citet{scheible-schulteimwalde-springorum:2013:IJCNLP} show that by restricting the contexts we use for building such distributional models, by carefully selecting context features such as by selecting verbs it is possible to overcome this problem to an extent.
Recall that 3-way co-occurrences require a third word co-occurring in the contexts that contain the co-occurrence between two words we are interested in measuring similarity. 
Therefore, 3-way embeddings by definition impose contextual restrictions that seem to be a promising alternative for pre-selecting contextual features. 
We plan to explore the possibility of using 3-way embeddings for discriminating antonyms in our future work.

\section{Conclusion}

We proved a theoretical relationship between joint probability of more than two words and their embeddings.
Next, we learnt word embeddings using $k$-way co-occurrences to understand the types of information captured in a $k$-way co-occurrence.
Our experimental results empirically validated the derived relationship.
Moreover, by considering $k$-way co-occurrences beyond $2$-way co-occurrences, we can learn better word embeddings for tasks that require contextual information such as analogy detection and short-text classification.

\section*{Appendix}

In this supplementary, we prove the main theorem (Theorem 1) stated in the paper.
The definitions of the symbols and notation are given in the paper.

\begin{theorem}
  Suppose the word vectors satisfy (2) in the paper. Then, we have
  \begin{align*}
    \log p(w_1,\ldots,w_k) = \frac{\norm{\sum_{i\in [k]} \vec{w}_{i}}_{2}^2}{2d} - k \log Z \pm \epsilon.
  \end{align*}
  for $\epsilon = O(k \epsilon_z) + \widetilde{O}(1/d) + O(k^2 \epsilon_2)$.
\end{theorem}
\begin{proof}

  Let $c_i\;(i \in \set{1,\ldots,k})$ be the hidden discourse that determines the probability of word $w_i$.
  We use $p(c_{i+1} \mid c_i)\;(i \in \set{1,\ldots,k-1})$ to denote the Markov kernel (transition matrix) of the Markov chain.
  Let $\caC$ be the stationary distribution of discourse $c$, and $\caD$ be the joint distribution of $c_{[k]} = (c_1,\ldots,c_k)$.
  We marginalise over the contexts $c_1,\ldots,c_k$ and then use the independence of $w_1,\ldots,w_k$ conditioned on $c_1,\ldots,c_k$,
  \begin{align}
    p(w_1,\ldots,w_k)
    = \Ep_{c_{[k]} \sim \caD}\Bigl[\prod_{i \in [k]} \frac{\exp(\langle \vec{w}_{i}, \vec{c}_i\rangle)}{Z_{c_i}}  \Bigr]. \label{eq:joint-probability-1}
  \end{align}

  We first get rid of the partition function $Z_{c_i}$ using Lemma~1 stated in the paper.
  Let $\caF_i$ be the event that $c_i$ satisfies
  \begin{align*}
    (1-\epsilon_z)Z \leq Z_{c_i} \leq (1+\epsilon_z)Z.
  \end{align*}
  Let $\caF = \bigcap_{i \in [k]} \caF_i$, and $\overline{\caF}$ be its negation.
  Moreover, let $\bmone_\caF$ be the indicator function for the event $\caF$.
  By Lemma~1 and the union bound, we have $\Ep[\bmone_\caF ] = \Pr[\caF ] \geq 1 - k\exp(-\Omega(\log^2 n))$.

  We first decompose~\eqref{eq:joint-probability-1} into the two parts according to whether event $\caF$ happens, that is,
\begin{align}
    p(w_1,\ldots,w_k)   =     \Ep_{c_{[k]} \sim \caD}\Bigl[\prod_{i \in [k]} \frac{\exp(\langle \vec{w}_{i}, \vec{c}_i\rangle)}{Z_{c_i}} \bmone_\caF \Bigr]
    +
    \Ep_{c_{[k]} \sim \caD}\Bigl[\prod_{i \in [k]} \frac{\exp(\langle \vec{w}_{i}, \vec{c}_i\rangle)}{Z_{c_i}} \bmone_{\overline{\caF}} \Bigr] \label{eq:split}
 \end{align}

  We bound the first quantity using (1) in the paper and the definition of $\caF$.
  That is,
  \begin{align}
 \Ep_{c_{[k]} \sim \caD}\Bigl[\prod_{i \in [k]} \frac{\exp(\langle \vec{w}_{i}, \vec{c}_i\rangle)}{Z_{c_i}} \bmone_\caF \Bigr]
  \leq
    \frac{(1+\epsilon_z)^k}{Z^k} \Ep_{c_{[k]} \sim \caD}\Bigl[\prod_{i \in [k]} \exp(\langle \vec{w}_{i}, \vec{c}_i\rangle) \bmone_\caF \Bigr].
    \label{eq:first-quantity}
  \end{align}

  For the second quantity, we have by Cauchy-Schwartz,
  \begin{align}
    \left(\Ep_{c_{[k]} \sim \caD}\Bigl[\prod_{i \in [k]} \frac{\exp(\langle \vec{w}_{i}, \vec{c}_i\rangle)}{Z_{c_i}} \bmone_{\overline{\caF}} \Bigr]\right)^2
    & \leq \prod_{i\in [k]} \left( \Ep_{c_{[k]} \sim \caD}\Bigl[ \frac{\exp(\langle \vec{w}_{i}, \vec{c}_i\rangle)^2}{Z_{c_i}^2} \bmone_{\overline{\caF}} \Bigr]  \right) \notag \\
 &    \leq
    \prod_{i\in [k]} \left( \Ep_{c_i}\Bigl[ \frac{\exp(\langle \vec{w}_{i}, \vec{c}_i\rangle)^2}{Z_{c_i}^2} \Ep_{c_{-i} \mid c_i}\bmone_{\overline{\caF}} \Bigr]  \right),
    \label{eq:second-quantity}
  \end{align}
  where $c_{-i}$ denotes the tuple obtained by removing $c_i$ from the tuple $(c_1,\ldots,c_k)$.

  Using the argument as in~\cite{Arora:TACL:2016}, we can show that
 \[
    \Ep_{c_i}\Bigl[ \frac{\exp(\langle \vec{w}_{i}, \vec{c}_i\rangle)^2}{Z_{c_i}^2} \Ep_{c_{-i} \mid c_i}\bmone_{\overline{\caF}}\Bigr] \leq k\exp(-\Omega(\log^{1.8} n)).
  \]
  Hence by~\eqref{eq:second-quantity}, the second quantity is bounded by $k^k\exp(-\Omega(k\log^{1.8 }n))$.
  Combining this with~\eqref{eq:split} and~\eqref{eq:first-quantity}, we obtain
  \begin{align*}
    p(w_1,\ldots,w_k) & \leq
    \frac{(1+\epsilon_z)^k}{Z^k} \Ep_{c_{[k]} \sim \caD}\Bigl[\prod_{i \in [k]} \exp(\langle \vec{w}_{i}, \vec{c}_i\rangle) \bmone_\caF \Bigr]
    + k^k\exp(-\Omega(k \log^{1.8} n)) \notag \\
  &   \leq
    \frac{(1+\epsilon_z)^k}{Z^k} \Ep_{c_{[k]} \sim \caD}\Bigl[\prod_{i \in [k]} \exp(\langle \vec{w}_{i}, \vec{c}_i\rangle) \Bigr] + \delta_0
  \end{align*}
  where
  \[ \delta_0 = k^k\exp(-\Omega(k\log^{1.8} n)) Z^k \leq k^k\exp(-\Omega(k\log^{1.8} n)).
  \]
  The last inequality is due to the fact that $Z \leq \exp(2\kappa)n = O(n)$, where $\kappa$ is the upper bound on $s$ used to generate word embedding vectors, which is regarded as a constant.

  On the other hand, we can lowerbound similarly
  \[
    p(w_1,\ldots,w_k)
    \geq
    \frac{(1-\epsilon_z)^k}{Z^k} \Ep_{c_{[k]} \sim \caD}\Bigl[\prod_{i \in [k]} \exp(\langle \vec{w}_{i}, \vec{c}_i\rangle) \Bigr] - \delta_0
  \]
  Taking logarithm, the multiplicative error translates to an additive error
  \begin{align*}
    \log p(w_1,\ldots,w_k) =& \log \left( \Ep_{c_{[k]} \sim \caD} \left[ \prod_{i \in [k]}\exp(\langle \vec{w}_{i}, \vec{c}_i\rangle) \right] \pm \delta_0 \right)
   - k\log Z +k\log(1 \pm \epsilon_z).
  \end{align*}
  For the purpose of exploiting the fact that $c_1,\ldots,c_k$ should be close to each other, we further rewrite $\log p(w_1,\ldots,w_k)$ by re-organizing the expectations above,
  \begin{align}
    \log p(w_1,\ldots,w_k) = \log \Bigl(A_1 \pm \delta_0 \Bigr)- k\log Z +k\log(1 \pm \epsilon_z).
    \label{eq:joint-probability-by-A}
  \end{align}
  where $A_1,\ldots,A_k$ are defined as
  \begin{align*}
    A_i &= \Ep_{c_i \mid c_1,\ldots,c_{i-1}} \Bigl[\exp(\langle \vec{w}_{i} , \vec{c}_i \rangle) A_{i+1}\Bigr] \quad i \in [k],\\
    A_{k+1} &= 1.
  \end{align*}
  Here, we regard $c_0$ is a discourse uniformly sampled from $\caC$.
  Then, we inductively show that
  \[ A_i = (1\pm \epsilon_2)^{k (k-i+1)}\exp \left( \langle \sum_{j=i}^{k} \vec{w}_{j}, \vec{c}_{i-1}\rangle \right). \]
  The base case $i = k+1$ clearly holds.

  Suppose that the claim holds for $i+1$.
  Since $\norm{\vec{w}_{j}} \leq \kappa\sqrt{d}$ for every $j \in \set{i,\ldots,k}$, we have that
  \begin{align*}
  \langle \sum_{j=i}^{k}\vec{w}_{j} , \vec{c}_{i-1} - \vec{c}_i\rangle \leq \norm{\sum_{j=i}^{k} \vec{w}_{j}} \norm{\vec{c}_{i-1} - \vec{c}_{i}} \leq k \kappa\sqrt{d} \norm{\vec{c}_{i-1} - \vec{c}_i}.
  \end{align*}
  Then we can bound $A_k$ by
  \begin{align*}
    A_i
    & = \Ep_{c_i \mid c_{[i-1]}} \Bigl[\exp(\langle \vec{w}_{i} , \vec{c}_i\rangle)A_{i+1}\Bigr] \\
    & = (1 + \epsilon_2)^{k(k-i)}\Ep_{c_i \mid c_{[i-1]}}\Bigl[\exp\Bigl(\langle \vec{w}_{i} + \sum_{j=i+1}^{k} \vec{w}_{j} , \vec{c}_i\rangle\Bigr) \Bigr] \\
    & = (1 + \epsilon_2)^{k(k-i)}\exp\Bigl(\langle \sum_{j=i}^{k} \vec{w}_{j} , \vec{c}_{i-1} \rangle\Bigr)  \Ep_{c_i \mid c_{[i-1]}} \Bigl[\exp\Bigl(\langle \sum_{j=i}^{k} \vec{w}_{j} , \vec{c}_i - \vec{c}_{i-1}\rangle\Bigr)\Bigr] \\
    & \leq (1 + \epsilon_2)^{k(k-i)} \exp\Bigl(\langle \sum_{j=i}^{k} \vec{w}_{j} , \vec{c}_{i-1}\rangle\Bigr) \Ep\Bigl[\exp(\kappa k \sqrt{d}\norm{ \vec{c}_{i-1} - \vec{c}_i })\Bigr] \\
    & \leq (1 + \epsilon_2)^{k(k-i+1)} \exp\Bigl(\langle \sum_{j=i}^k\vec{w}_{j} , \vec{c}_{i-1}\rangle\Bigr),
  \end{align*}
  where the last inequality follows from our model assumptions.

  To derive a lower bound on $A_i$, observe that
 \begin{align*}
    \Ep_{c_i \mid c_{[i-1]}}\Bigl[\exp(\kappa \sqrt{d}\norm{\vec{c}_{i-1}- \vec{c}_i})\Bigr]+ \Ep_{c_i \mid c_{[i-1]}}\Bigl[\exp(-\kappa \sqrt{d}\norm{\vec{c}_{i-1}-\vec{c}_i})\Bigr]\geq 2.
  \end{align*}
  Therefore, our model assumptions imply that
  \[
    \Ep_{c_i\mid c_{[i-1]}}[\exp(-\kappa \sqrt{d}\norm{\vec{c}_{i-1}-\vec{c}_i})]\geq 1-\epsilon_2
  \]
  Hence by induction,
  \begin{align*}
    A_i &= (1-\epsilon_2)^{k(k-i)} \exp\Bigl(\langle \sum_{j = i}^k \vec{w}_{j} , \vec{c}_{i-1}\rangle\Bigr) \Ep_{c_i \mid c_{[i-1]}} \exp\Bigl(\langle \sum_{j = i}^k \vec{w}_{j} , \vec{c}_i - \vec{c}_{i-1}\rangle\Bigr) \\
    &\geq (1 - \epsilon_2)^{k(k-i+1)} \exp\Bigl(\langle \sum_{j = i}^k \vec{w}_{j} , \vec{c}_{i-1}\rangle\Bigr).
  \end{align*}
  Therefore, we obtain that
  \[
  A_1 = (1 \pm  \epsilon_2)^{k^2}\exp\Bigl(\langle \sum_{i=1}^k \vec{w}_{i}, \vec{c}_0 \rangle\Bigr).
  \]

  Plugging the just obtained estimate of $A_1$ into~\eqref{eq:joint-probability-by-A}, we get
  \begin{align}
    \log p(w_1,\ldots,w_k) =    \log \Bigl( \Ep_{c \sim \caC}\Bigl[\langle \sum_{i=1}^k \vec{w}_{i}, \vec{c} \rangle\Bigr] \pm \delta_0 \Bigr) - k\log Z + k\log(1\pm \epsilon_z) + k^2\log(1 \pm \epsilon_2).
    \label{eq:joint-probability-by-inner-product}
  \end{align}
  Now it suffices to compute $\E_c[\exp(\langle \sum_{i=1}^k\vec{w}_{i}, \vec{c}\rangle)]$.
  Note that if $c$ had the distribution $\caN(0, I/d)$, which is very similar to uniform distribution over the sphere, then we could get straightforwardly
   \begin{equation*}
    \Ep_c\left[ \exp \left( \Langle \sum_{i=1}^k\vec{w}_{i}, \vec{c} \Rangle \right) \right] = \exp \left( \frac{\norm{\sum_{i=1}^k\vec{w}_{i}}^2} {2d} \right).
  \end{equation*}
  For $\vec{c}$ having a uniform distribution over the sphere, by Lemma A.5 in~\cite{Arora:TACL:2016},
  the same equality holds approximately,
  \[
    \sum_{c}\left[ \exp \left( \Langle \sum_{i=1}^k \vec{w}_{i}, \vec{c}\Rangle \right) \right] = (1 \pm \epsilon_3)\exp\left( \frac{\norm{\sum_{i=1}^k \vec{w}_{i}}^2}{2d}\right).
  \]
  where $\epsilon_3 = O (1/d)$.
  Plugging this into~\eqref{eq:joint-probability-by-inner-product}, we have that
  \begin{align*}
     \log p(w_1,\ldots,w_k) = &\log \left((1 \pm \epsilon_3)\exp\left( \frac{\norm{\sum_{i=1}^k \vec{w}_{i}}_2}{2d}\right) \pm \delta_0\right)
    -k\log Z+k\log(1 \pm \epsilon_z)+k^2\log(1 \pm \epsilon_2) \\
     =&  \frac{\|\sum_{i=1}^k \vec{w}_{i}\|_2}{2d}+O(\epsilon_3)+O(\delta_0') -k\log Z \pm k \epsilon_z  \pm k^2 \epsilon_2
  \end{align*}
  where
  \begin{align*}
  \delta_0' = \delta_0 \cdot \left(\Ep_{c \sim \caC} \left[\exp\left(\Langle \sum_{i=1}^k \vec{w}_{i}, \vec{c}\Rangle \right)\right]\right) = k^k\exp(-\Omega(k\log^{1.8} n)).
  \end{align*}

  Note that $\epsilon_3 = \widetilde{O}(1/d)$, $\epsilon_z = \widetilde{O} (1/ \sqrt{n})$ by assumption.
  Therefore, we obtain that
  \begin{align*}
   \log p(w_1,\ldots,w_k) =  \frac{\norm{\sum_{i=1}^k \vec{w}_{i}}^2}{2d} - k \log Z \pm O(k\epsilon_z)+O(k^2\epsilon_2)+O (1/d).
   & \qedhere
  \end{align*}
\end{proof}

\bibliography{sentemb.bib}
\bibliographystyle{plainnat}

\end{document}